\documentclass[final,conference]{IEEEtran}
\usepackage{cite}
\usepackage{amsmath,amssymb,amsfonts}
\usepackage{algorithmic}
\usepackage{graphicx}
\usepackage{textcomp}
\usepackage{xcolor}
\def\BibTeX{{\rm B\kern-.05em{\sc i\kern-.025em b}\kern-.08em
    T\kern-.1667em\lower.7ex\hbox{E}\kern-.125emX}}

\usepackage{amsthm}    
\usepackage[utf8]{inputenc}
\usepackage{booktabs}
\usepackage{multirow}
\usepackage{comment}
\usepackage[ruled,vlined]{algorithm2e}
\usepackage[colorlinks]{hyperref}

\ifCLASSOPTIONcompsoc
\usepackage[caption=false,font=normalsize,labelfont=sf,textfont=sf]{subfig}
\else
\usepackage[caption=false,font=footnotesize]{subfig}
\fi

\newcommand{\task}{\mathcal{T}}
\newcommand{\expecwrt}[2]{\mathbb{E}_{#1}{#2}}
\newcommand{\three}{.30}
\newcommand{\four}{.22}

\title{Compressed Hierarchical Representations for Multi-Task Learning and Task Clustering
}
\author{\IEEEauthorblockN{João Machado de Freitas\IEEEauthorrefmark{1}, Sebastian Berg\IEEEauthorrefmark{2}, Bernhard C. Geiger\IEEEauthorrefmark{1} and Manfred Mücke\IEEEauthorrefmark{2}}
\IEEEauthorblockA{\IEEEauthorrefmark{1}Know-Center GmbH, Graz, Austria
}
\IEEEauthorblockA{\IEEEauthorrefmark{2}Materials Center Leoben Forschung GmbH, Leoben, Austria\\
Email: \{jfreitas,geiger\}@ieee.org, \{sebastian.berg,manfred.muecke\}@mcl.at}
}
\begin{document}
\bstctlcite{IEEEexample:BSTcontrol}  % et al

\maketitle
\thispagestyle{plain}
\pagestyle{plain}

\begin{abstract}
In this paper, we frame homogeneous-feature multi-task learning (MTL) as a hierarchical representation learning problem, with one task-agnostic and multiple task-specific latent representations. Drawing inspiration from the information bottleneck principle and assuming an additive independent noise model between the task-agnostic and task-specific latent representations, we limit the information contained in each task-specific representation.

It is shown that our resulting representations yield competitive performance for several MTL benchmarks. Furthermore, for certain setups, we show that the trained parameters of the additive noise model are closely related to the similarity of different tasks. This indicates that our approach yields a task-agnostic representation that is disentangled in the sense that its individual dimensions may be interpretable from a task-specific perspective.
\end{abstract}


\section{Introduction}
Multi-task learning (MTL) is a growing field of research that aims to utilize synergies between different tasks to reduce the amount of data or computational resources required for these tasks. Applications of MTL are manifold and range over areas as diverse as natural language processing (e.g., the accuracy of semantic role labeling improves if learned jointly with different tasks, such as part-of-speech tagging or named entity recognition~\cite{collobert08unified}); computer vision (e.g., by jointly learning object detection and object classification~\cite{RenHG015}); or autonomous driving (e.g., simultaneous detection of static and dynamic objects~\cite{ishihara2021multi}).

In this work, we frame homogeneous-feature MTL (where each task relies on the same set of features~\cite{Zhang_Survey}) as a hierarchical representation learning problem, combining the ideas from~\cite{nautrup2020operationally}, which restricts the information different tasks can access, and~\cite{qian2021mvib}, which proposes an information-theoretic approach to representation-based MTL.
In the context of MTL in deep neural networks, our approach uses hard parameter sharing, opposed to soft parameter sharing~\cite{ruder2017overview}.
Specifically, we design a single task-agnostic representation $Z$, from which several task-specific representations $W_j$ are derived via adding a Gaussian noise vector with diagonal covariance matrix. In other words, we assume an additive, independent (but not identically distributed) noise model between the task-agnostic and the task-specific representations, thus resulting in a hierarchical model. Making the noise variances trainable and employing an optimization objective motivated by the information bottleneck (IB) principle~\cite{Tishby_InformationBottleneck}, our task-specific representations are \emph{compressed}, while the task-agnostic representation remains unconstrained (in contrast to~\cite{qian2021mvib}). While the individual dimensions of our task-agnostic representation $Z$ are not as readily interpreted as in the case of~\cite{nautrup2020operationally}, we show that -- for an appropriate selection of tasks and adequate hyper-parameters -- the noise vectors defining the task-specific representations can be used to cluster tasks based on their similarity. In summary, our contribution is four-fold: We
\begin{itemize}
    \item propose a variational optimization objective for hierarchical representation learning with compressed task-specific representations (Section~\ref{sec:method});
    \item theoretically analyze the objective for tasks described by jointly Gaussian random variables (Section~\ref{sec:analysis});
    \item show that our objective yields competitive performance across several MTL benchmarks, including MTFL and MultiMNIST (Section~\ref{sec:benchmarks});
    \item and illustrate on synthetic and realistic examples that our approach can be used to assess task similarity (Section~\ref{sec:taskclustering}).
\end{itemize}
We discuss benefits, future work and possible limitations in Section~\ref{sec:discussion}.

\section{Related Work}
\label{sec:related}
Our work mostly draws from two sources: (i) it is inspired by information-theoretic approaches to machine learning and MTL and (ii) it has strong connections to task clustering based on learned representations.

In the former field, our work is inspired by the information-theoretic MTL framework proposed by~\cite{qian2021mvib}. There, the authors learn a single task-agnostic, compressed representation $Z$ that is useful for all downstream tasks (compression is deemed necessary to ensure positive information transfer between tasks, cf.~\cite{Wu2020Understanding}). This representation is obtained by optimizing a generalization of the variational IB~\cite{Alemi_DVIB}, weighing multiple tasks w.r.t.\ each other according to their respective uncertainties, cf.~\cite{Kendall_2018_CVPR}. A similar setup was studied in~\cite{Vera_MTL}, where the authors show that -- for a fixed vector of task weights -- MTL with cross-entropy risk is equivalent to an IB problem with side information at the decoder. The authors propose an iterative algorithm to solve this latter IB problem~\cite[Alg.~1]{Vera_MTL}.

As a second source of inspiration, our work builds on literature that clusters tasks, or evaluates task similarity, based on trained parameters of MTL models, cf.~\cite[Sec.~2.3 \& 2.4]{Zhang_Survey}. This line of research is motivated by the fact that the choice of which tasks to combine will influence the overall performance of the system -- something recently investigated for computer vision by~\cite{standley20which}. A classical example of such work is~\cite{Kumar_MTL}, which assumes that the parameters of the (linear) task-specific models are obtained as sparse linear combinations of a small number of latent task parameter vectors; using tensor instead of matrix factorization, this method was extended to neural networks in~\cite{Yang_MTL}. Moreover, it can be shown that the resulting sparsity patterns give insights into the similarity between different tasks, e.g.,~\cite[Figs.~3 \& 4]{Kumar_MTL} or~\cite[Fig.~4]{Yang_MTL}. In a similar direction, the authors of~\cite{Shen_Gumbel} formulate a Bayesian framework for MTL and define the prior distributions for the latent representation and classifier for a selected task as a linear combination of the corresponding variational posteriors of all other tasks. They show that the coefficients of this linear combination yield insights about task similarity, cf.~\cite[Fig.~3]{Shen_Gumbel}.
The authors of~\cite{AAAI2021} propose a loss function that depends on (learned) similarities between the task-dependent class-conditional data distributions, which improves robustness to distribution shifts.
In~\cite{Wu2020Understanding}, the authors measure task similarity via approximating the covariance between the task-specific input data. They use this approximation to (linearly) align the inputs to facilitate learning a task-agnostic representation.

Task similarity, which gives rise to similarities between model parameters, in turn, has the potential to result in similarities in latent representations. This connects this branch of literature to the field of disentangled representation learning, where latent representations $Z$ with individual dimensions interpretable using human-interpretable concepts are preferred~\cite{bengio2013representation}. A work in this direction is~\cite{nautrup2020operationally}, where the authors convert data from reference experiments to representations $Z$ that are physically meaningful. They achieve this by using these representations to predict results of other experiments, and by constraining the dimensions of $Z$ that the respective prediction may utilize.

Our work is connected to the related work as follows: Just as~\cite{nautrup2020operationally,qian2021mvib,Vera_MTL}, we consider the setting where different tasks operate on the same input features, i.e., we consider homogeneous-feature MTL. While~\cite{qian2021mvib} proposes a single compressed task-agnostic representation $Z$, we do not constrain the task-agnostic, but only the separate, task-specific representations $\{W_j\}$; we obtain these task-specific representations by adding independent noise to $Z$, an approach that is inspired by~\cite{nautrup2020operationally}. Similar to~\cite{Kumar_MTL,Yang_MTL}, we utilize parameters of the learned MTL model -- namely, the properties of said independent noise vectors -- to infer the similarity of different tasks. Finally, while~\cite{qian2021mvib} trades between different tasks using their inherent uncertainty, we do not weigh different tasks differently, but leave this for future work.

\section{Hierarchical Representations for Multi-Task Learning}
\label{sec:method}
\begin{figure}[htbp]
    \centering
    \includegraphics[
        trim=0 12 0 12,
        clip,
        width=.48\textwidth
    ]{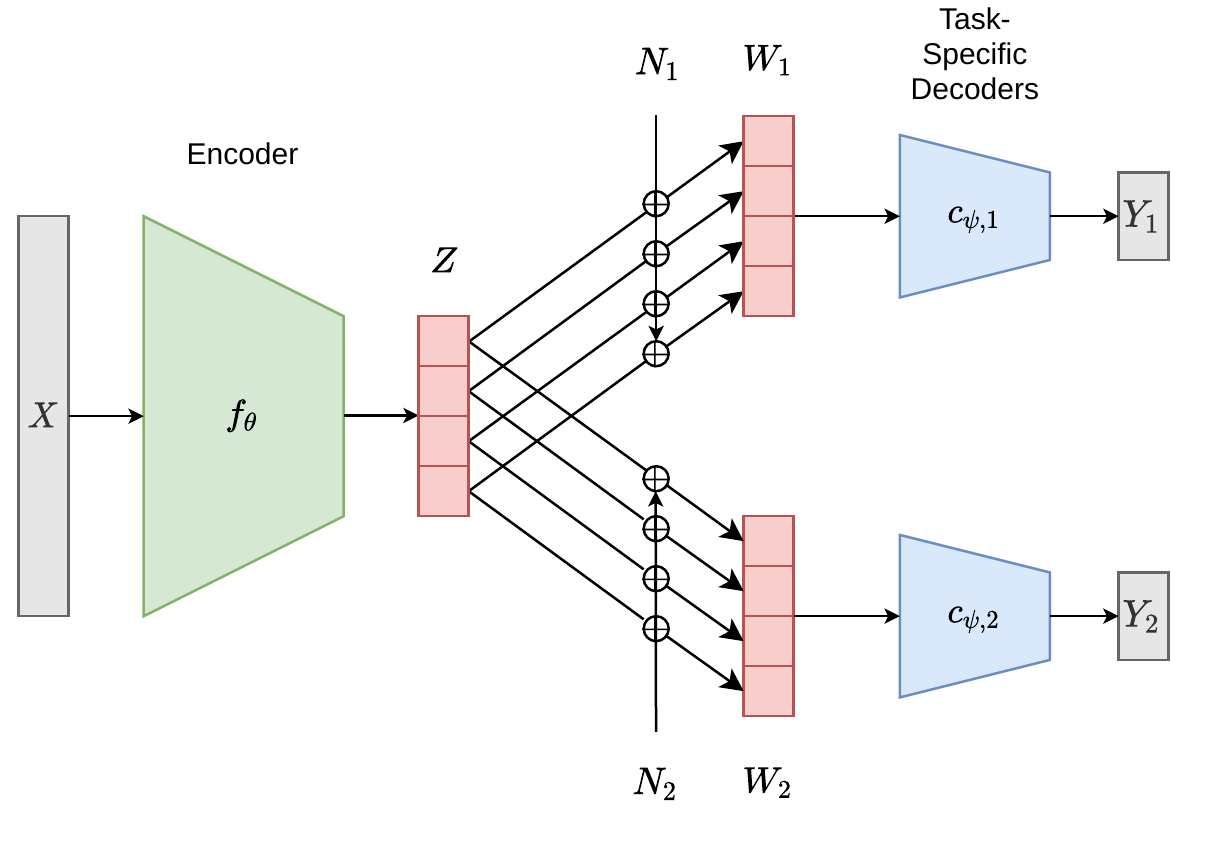}
    \caption{Hierarchical representations for multi-task learning. Our aim is to learn a common, task-agnostic representation $Z$ and multiple task-specific representations $W_j$ that are regularized by the information bottleneck principle. See text for details.}
    \label{fig:structure}
\end{figure}

We consider a homogeneous-feature MTL setting with $L$ tasks $\task_1,\dots,\task_L$. Each task is characterized by a conditional distribution $p_{Y_j|X}$, where we call $X$ the feature and $Y_j$ the target of task $\task_j$. We further assume that we have access to a dataset $\mathcal{D}=\{(x_n,y_{1,n},\dots,y_{L,n})\}_{n=1,\dots,N}$ drawn i.i.d.\ from $p_{X,Y_1,\dots,Y_L}=p_X\prod_{j=1}^L p_{Y_j|X}$, where $p_X$ is the marginal distribution of the features. For the sake of exposition we further suppose that $\task_j$ are classification tasks, i.e., $Y_j$ are discrete random variables. The extension to regression tasks is easily possible.

\subsection{Model}
We aim to create a hierarchical set of latent representations for multi-task learning, namely a \emph{task-agnostic} representation that is unconstrained, and several \emph{task-specific} representations that are regularized to be concise in the sense of a minimum sufficient statistic. More specifically, we propose the model depicted in \figurename~\ref{fig:structure}. 

In this model, the $D$-dimensional task-agnostic representation $Z$ is obtained via encoding the feature $X$ with a deterministic encoder $f_\theta$, parameterized by $\theta$; i.e., $Z=f_\theta(X)$. The $D$-dimensional task-specific representations are obtained from $Z$ by adding zero-mean Gaussian noise with diagonal covariance matrix, i.e., for every task $\task_j$ we have
\begin{equation}
    W_j=Z+N_j,\quad N_j\sim\mathcal{N}(0,\Sigma_j)
\end{equation}
where $\Sigma_j$ is diagonal with entries $\sigma^2_{j,i}>0$, $i=1,\dots, D$.

\subsection{Lagrangian of the Learning Problem}
The task-specific representations $W_j$ should simultaneously allow solving the task $\task_j$ and be concise in the sense that they do not contain more information than necessary for this task. In other words, $W_j$ should be minimum sufficient statistics of $X$ for $Y_j$. This immediately suggests the IB principle~\cite{Tishby_InformationBottleneck} for learning these representations, which aims at solving
\begin{equation}\label{eq:cost_function}
    \min_{\theta,\{\Sigma_j\}}\ \sum_{j=1}^L H(Y_j|W_j) + \beta_j I(X;W_j).
\end{equation}
where $\beta_j$ is a task-specific parameter trading between sufficiency and conciseness. Intuitively, the encoder parameters $\theta$ are chosen such that $Z$ is sufficient for all tasks, while the noise variances $\Sigma_j$ are chosen such that the representations $W_j$ are as concise as possible. Due to the diagonal covariance structure of $\Sigma_j$, the encoder is further incentivized to disentangle the dimensions of $Z$ such that subsequent tasks can rely on small subsets of $Z$'s dimensions, cf.~\cite{nautrup2020operationally}. 

\subsection{Variational Bounds}
We now propose variational bounds on the terms in~\eqref{eq:cost_function}. Our bounds are closely related to those in~\cite{Alemi_DVIB}.

\subsubsection{Minimizing $H(Y_j|W_j)$}
Since the true posterior $p_{Y_j|W_j}$ is not available, we use the standard variational bound obtained by training a neural classifier $c_{\psi,j}$ with parameters $\psi$ for $Y_j$ that takes $W_j$ as input. Letting $c_{\psi,j}(y|w)$ denote the estimated posterior probability of class $y$ for latent representation $w$, and since cross-entropy bounds entropy from above, we get
\begin{align}\label{eq:crossentropy}
 H(Y_j|W_j) &\le -\expecwrt{W_j}{\sum_{y} p_{Y_j|W_j}(y|W_j)\log c_{\psi,j}(y|W_j)}.
\end{align}
\subsubsection{Minimizing $I(X;W_j)$}
For the second term, $I(X;W_j)$, we use the usual variational bound. Specifically, let $q_j$ be an arbitrary distribution on the $D$-dimensional latent space. Thus,
\begin{align}
    & I(X;W_j)\notag\\
    &\ =\expecwrt{X,W_j}{\log{\frac{p_{W_j|X}(W_j|X)}{p_{W_j}(W_j)}}}\\
    &\ =\expecwrt{X,W_j}{\log{\frac{p_{W_j|X}(W_j|X)q_j(W_j)}{p_{W_j}(W_j)q_j(W_j)}}}\\
    &\ =\expecwrt{X}{D(p_{W_j|X}(\cdot|X)\Vert q_j(\cdot))}+ D(p_{W_j}\Vert q_j)\\
    &\ \le \expecwrt{X}{D(p_{W_j|X}(\cdot|X)\Vert q_j(\cdot))}\label{eq:varbound}
\end{align}
where $D$ denotes the Kullback-Leibler divergence.

Since this holds for any marginal distribution $q_j$, we can minimize the r.h.s. of~\eqref{eq:varbound} within a family of distributions to improve the bound. Specifically, let $q_j$ denote a Gaussian distribution with learnable mean $\nu_j$ and learnable diagonal covariance matrix $\Xi_j$ with entries $\xi^2_{j,i}$, $i=1,\dots, D$. Since, by our model, $W_j$ is Gaussian conditioned on $X$, we can express the Kullback-Leibler divergence in~\eqref{eq:varbound} in closed form and bound $I(X;W_j)$ by minimizing
\begin{multline}
 \expecwrt{X}{\sum_{i=1}^D 
    \left(
        \log\left(\frac{\xi_{j,i}^2}{\sigma^2_{j,i}}\right)
        + \frac{\sigma^2_{j,i} + (\nu_{j,i} - f_\theta(X))^2}{2\xi_{j,i}^2} - \frac{1}{2}
    \right)}
\end{multline}
over $\nu_j$ and $\Xi_j$.

\subsection{Implementation}
Our model is implemented with amortized inference of $W_j$, which allows to exploit the similarity between inputs to efficiently encode the underlying observations while also allowing to jointly optimize the parameters and representations for the whole dataset. In practice, $f_\theta$ is parameterized by a single NN and the $q_j$ and noise parameters are shared across all dataset observations.
Furthermore, the expectations in the objective are approximated by Monte Carlo sampling from the empirical distribution and the posterior. At train and test time, we only used one sample from the posterior.
Unsurprisingly, we also found that learning the variational means $\nu_{j,i}$ is unnecessary and training is stabler when $\nu_{j,i}=0$ is frozen.
Since $Z$ is a deterministic function of $X$, the modeling choice $Z+N_j$ is equivalent to applying the reparameterization trick~\cite{kingma2014autoencoding} to $W_j$, where $N_j=\varepsilon \odot \sigma_{i,j}$, with $\varepsilon \sim \mathcal N(0, I)$. Thus, we can perform gradient back-propagation through $W_j$.

While our Lagrangian formulation~\eqref{eq:cost_function} allows to select a different trade-off between sufficiency and conciseness for each task, for the sake of simplicity we chose the same $\beta$ for every task (i.e., we set $\beta_j=\beta$). This simplifies sensitivity and ablation analyses and establishes a lower bound on the performance of the proposed method.

\section{Theoretical Analysis}
\label{sec:analysis}

In our experiments in Section~\ref{sec:taskclustering} we observed that the variances of the noise vectors $N_j$ allow us to make statements about the similarity of the respective tasks. At the heart of this claim lies our assumption that \emph{distinct tasks will occupy disjoint sets of dimensions of the task-agnostic representation $Z$} (at least if its dimension $D$ is sufficiently large). We will now show that this assumption holds in an analytically tractable toy setting in which features $X$ and task target $Y_j$ are jointly Gaussian. While this assumption is practically unrealistic, we believe that the following analysis will nevertheless shed some light on the intuition behind our model.

For just a single task $Y$ in this Gaussian case, the representation $Z$ minimizing the IB functional
\begin{equation}\label{eq:IBGauss}
    -I(Y;Z) + \beta I(X;Z)
\end{equation}
is given by a noisy linear transform, i.e., $Z=AX+N$, where $Z$ is assumed to have the same dimension $D$ as $X$~\cite{Chechik_GaussIB}. It was observed that, for $N$ having an identity covariance matrix, the rank of $A$ reduces with increasing $\beta$, i.e., with stronger regularization, effectively reducing the dimensionality of $Z$~\cite[Th.~3.1]{Chechik_GaussIB}. Utilizing~\cite[Lemma~A.1]{Chechik_GaussIB}, we instead let $A$ remain full rank and vary the variances in $N$, letting some of them grow to infinity as $\beta$ increases.

Specifically, let $C_X$ be the covariance matrix of $X$ and $C_{X|Y}$ the conditional covariance matrix of $X$ given $Y$. Then, the IB functional~\eqref{eq:IBGauss} is also minimized by the representation $Z=AX+N$, where $A$ is the matrix of left-singular vectors $v_i$ of $C_{X|Y}C_X^{-1}$, ordered by the descending singular values $\lambda_i$. Further, $N\sim\mathcal{N}(0,\mathrm{diag}(\sigma_i^2))$, with
\begin{equation}\label{eq:noise_variances}
   \frac{1}{\sigma_i^2} = \alpha_i = \frac{\max\{0, \frac{1-\lambda_i}{\beta}-1\}}{\lambda_i v_i^T C_X v_i}
\end{equation}
and where $\beta$ is the Lagrangian parameter of the IB functional in~\eqref{eq:IBGauss}. Indeed, if $\beta>1/(1-\lambda_i)$, then $\alpha_\ell=0$ and $\sigma_\ell^2=\infty$ for all $\ell\ge i$.

Let us now return to the MTL setting for two tasks with targets $Y_1$ and $Y_2$, jointly Gaussian with $X$. Let us suppose further that $\beta_1>1/(1-\lambda_{1,D/2})$ and $\beta_2>1/(1-\lambda_{2,D/2})$, where $\lambda_{1,i}$ and $\lambda_{2,i}$ are the singular values of  $C_{X|Y_1}C_X^{-1}$ and  $C_{X|Y_2}C_X^{-1}$, respectively. Thus, the optimal representations for the separate IB problems rely only on at most a $D/2$-dimensional projection of $X$. This allows us to solve the MTL problem  to~\eqref{eq:cost_function} as follows: We obtain $A$ by stacking the dominant $D/2$ left-singular vectors of $C_{X|Y_1}C_X^{-1}$ and the dominant $D/2$ left-singular vectors of $C_{X|Y_2}C_X^{-1}$, respectively, rendering $Z=AX$ $D$-dimensional. We then construct $W_j=Z+N_j$, where the variances $\sigma_{j,i}^2$ are given by~\eqref{eq:noise_variances} for $Y=Y_j$, $j=1,2$. From this and our assumption on $\beta$ follows that of $N_1$ the last $D/2$ and of $N_2$ the first $D/2$ dimensions have infinite variance. In other words, in the task-agnostic representation $Z$ the two tasks are disentangled. 

This toy example required that the dimension $D$ of the latent representation and the regularization $\beta$ are sufficiently large to allow full disentanglement. Both $\beta$ and $D$ can be reduced if the tasks $\mathcal{T}_1$ and $\mathcal{T}_2$ are similar in the sense that some of the dominant left-singular vectors of $C_{X|Y_1}C_X^{-1}$ and $C_{X|Y_2}C_X^{-1}$ are identical. Then, the same dimension of $Z$ can serve both tasks. Similarly, for more than two tasks, task similarity can in this setting be intuitive interpreted as similarity of the (dominant) left-singular vectors of the $C_{X|Y_j}C_X^{-1}$: If two tasks share a higher number of identical dominant left-singular vectors than they do with the remaining tasks, then the respective noise variance vectors will have a higher number of small entries at identical positions.

\section{Experiments}

We perform two types of experiments. First, we evaluate our proposed method on benchmark MTL datasets and show that it performs competitively with similar representation-based MTL frameworks (Section~\ref{sec:benchmarks}). Second, we investigate under what conditions, and to what extent our proposed method can yield insights into the similarity of tasks (Section~\ref{sec:taskclustering}). In both experimental settings, we accompany our results with sensitivity and ablation analyses.

Our experiments are implemented in PyTorch~\cite{paszke2019pytorch}. A Jupyter notebook~\cite{kluyver2016jupyter} demonstration of our experiments, together with the related code is available\footnote{\url{https://github.com/jmachadofreitas/multitask-ijcnn2022}}.

\subsection{MTL Benchmarks}
\label{sec:benchmarks}

\textbf{Multi-Task Facial Landmark (MTFL).} For comparing model accuracy to state-of-the-art, we use the Multi-Task Facial Landmark (MTFL) dataset~\cite{zhang2014facial}. This dataset consists of two subsets of images showing faces with different resolutions.
The training subset contains 10,000 images, the evaluation   subset contains 2,995 images.
Four classification tasks are provided: (i) gender classification (male/female), (ii) smiling (yes/no), (iii) presence of glasses (yes/no), and (iv) head pose classification into five classes. All images were transformed to $256\times256$ pixels. 

Inspired by~\cite{qian2021mvib}, we used AlexNet~\cite{alex2014one} to solve the MTFL classification problems. All tasks share the encoder network down to the latent space layer. The remaining architectural detail are described in Table~\ref{tab:mtfl_arch}. For this experiment we used pre-trained weights up to the adaptive average pooling layer in Table~\ref{tab:mtfl_arch}. In addition, we used Adam optimizer~\cite{kingma2017adam} with learning rate $10^{-4}$, and we trained for 100 epochs. The noise variances of our information filters were initialized following a Gaussian distribution with a standard deviation of 0.01.  
We trained our models with varying $\beta$ values ($10^{-4}$, $10^{-3}$, $10^{-2}$, $10^{-1}$), as well as without adding noise at all, i.e., for $W_j=Z$. 
The parameter $\beta$ controls how much information the task-specific classifiers receive, with higher $\beta$ values restricting the information further.

We compare to our own implementation of the multi-task VIB (MTVIB) approach proposed in~\cite{qian2021mvib}. Most architectural and training details were identical to our model. The main difference is that MTVIB uses uncertainty weighting for the task losses and has no task-specific information filters, while our own model only relies on the information filters to modify the representations for each task.

\begin{table}[!htbp]
\renewcommand{\arraystretch}{1.3}
\caption{AlexNet Used for the MTFL Dataset}
\label{tab:mtfl_arch}
\centering
\begin{tabular}{c} 
    \hline
    \textbf{Encoder $f_\theta$}\\
    \hline
    Conv 64 (kernel 11x11, stride 4, padding 2) + ReLU\\ 
    MaxPool (kernel 3x3, stride 2) \\ 
    Conv 192 (kernel 5x5, stride 2, padding 2) + ReLU\\
    MaxPool (kernel 3x3, stride 2) \\ 
    Conv 384 (kernel 3x3, stride 2, padding 1) + ReLU\\
    Conv 256 (kernel 3x3, stride 2, padding 1) + ReLU\\
    Conv 256 (kernel 3x3, stride 2, padding 1) + ReLU\\
    MaxPool (kernel 3x3, stride 2) + AdaptiveAvgPool(input 6x6) \\ 
    Flattening + Linear (9216 $\rightarrow$ 256)\\
    \hline
    \textbf{Adding Noise $\{\sigma_{j,i}^2\}$}\\
    \hline
    \textbf{Task-Specific Classifiers $c_{\psi,j}$}\\
    Dropout (0.5) +
    Linear (256 $\rightarrow$ 256) + ReLU\\
    Linear (256 $\rightarrow$ out) + Softmax \\
    \hline
\end{tabular}
\end{table}
\begin{table}[!htbp]
\renewcommand{\arraystretch}{1.3}
\caption{Architecture for the Multi-MNIST and Grouped-MNIST Datasets
}
\label{tab:mnist_arch}
\centering
\begin{tabular}{ c } 
    \hline
    \textbf{Encoder $f_\theta$} \\
    \hline
    Flattening \\
    Linear (784 $\rightarrow$ 32) + ReLU \\
    Linear (32 $\rightarrow$ 32) + ReLU \\
    Linear (32 $\rightarrow$ 4) \\
 \hline
 \textbf{Adding Noise $\{\sigma_{j,i}^2\}$}\\
 \hline
 \textbf{Task-Specific Classifiers $c_{\psi,j}$}\\
 \hline
    Linear (4 $\rightarrow$ out) + Softmax \\
    \hline
\end{tabular}
\end{table}

\textbf{MultiMNIST.} We also compared the model accuracy on MultiMNIST dataset~\cite{sabour2017dynamic}, which consists on classifying two partly overlayed digits on the same image. The architecture is described in Table~\ref{tab:mnist_arch} and the training conditions are similar to those for MTFL, for both MTVIB and our model.

\textbf{Sensitivity and Ablation Analysis.} 
The results of the MTFL and MultiMNIST experiments with and without noise can be seen in Tables~\ref{tab:mtfl_res} and~\ref{tab:multimnist_res}, respectively.
The first row ($\beta$: '-') indicates that no noise was added, while the second row ($\beta$: '0') refers to adding noise but not penalizing information flow between task-agnostic and task-specific representations.

Regarding Table~\ref{tab:mtfl_res}, with small enough $\beta$, the model achieves an accuracy comparable to the ablated setting, where no noise is added to the task-agnostic representation. It can be seen that mild regularization ($\beta=10^{-4}$) improves generalization performance when compared to unregularized setting.

Note that the $\beta$ values for MTVIB and our method are not directly comparable. However, looking at the range it can be seen that the models perform comparably.
For increasing $\beta$, we observe a drop of accuracy, e.g., for $\beta$ between $0.01$ and $0.1$ for Gender classification. In this setting, the task-specific classifiers cannot access sufficient information from $Z$.

Concerning the results on the MultiMNIST dataset (Table~\ref{tab:multimnist_res}), we see that setting $\beta=0$ while adding noise allows to achieve a comparable performance to the MTVIB. It can also be seen that for positive $\beta$ our model seems to be less sensitive to $\beta$ than MTVIB, indicating that performance-limiting information filtering in our method may occur only at larger $\beta$ values.
\begin{table}[ht]
\renewcommand{\arraystretch}{1.3}
\caption{Accuracy (\%) on the MTFL Dataset}
\label{tab:mtfl_res}
\centering
\begin{tabular}{ cccccc } 
\hline
\bfseries Model                 & $\pmb\beta$  & \bfseries Gender & \bfseries Smile & \bfseries Glasses & \bfseries Head Pose \\
\hline
No Noise             & -         & 78.4  & 68.9  & 95.2  &  68.6 \\
\hline
\multirow{5}{*}{Ours} & 0         & 77.8  & 65.6  & 93.3  & 69.9 \\
                      & $10^{-4}$ & 75.1  & 66.4  & 94.3  & 69.6 \\
                      & $10^{-3}$ & 75.1  & 65.9  & 93.3  & 67.9 \\
                      & $10^{-2}$ & 71.5  & 63.4  & 91.1  & 63.9 \\
                      & $10^{-1}$ & 55.0  & 52.9  & 90.4  & 59.8 \\
\hline
\multirow{5}{*}{MTVIB} & 0         & 73.4  & 69.6  & 95.3  & 69.7  \\
                       & $10^{-4}$ & 77.3  & 70.1  & 94.5  & 69.6  \\
                       & $10^{-3}$ & 72.6  & 68.1  & 93.8  & 67.3  \\
                       & $10^{-2}$ & 70.8  & 64.1  & 91.9  & 60.2  \\
                       & $10^{-1}$ & 40.8  & 52.2  & 90.4  & 59.8  \\
\hline
\end{tabular}
\end{table}
\begin{table}[htbp]
\renewcommand{\arraystretch}{1.3}
\caption{\label{tab:multimnist_res}Accuracy (\%) on the MultiMNIST Dataset}
\centering
\begin{tabular}{ cccc } 
\hline
\bfseries Model       & $\pmb\beta$   & \bfseries Digit 1 & \bfseries Digit 2 \\
\hline
No Noise              & -         & 87.9 & 86.8 \\
\hline
\multirow{5}{*}{Ours}  
                    & 0         & 87.3 & 86.6 \\
                    & $10^{-4}$ & 87.6 & 87.0 \\
                    & $10^{-3}$ & 87.4 & 86.8 \\
                    & $10^{-2}$ & 86.9 & 87.0 \\
                    & $10^{-1}$ & 85.6 & 85.9 \\
\hline
\multirow{5}{*}{MTVIB} & 0      & 87.1 & 86.6  \\
                    & $10^{-4}$ & 86.7 & 85.7 \\
                    & $10^{-3}$ & 83.1 & 81.2 \\
                    & $10^{-2}$ & 80.5 & 77.9 \\
                    & $10^{-1}$ & 80.5 & 77.9 \\
\hline
\end{tabular}
\end{table}
\subsection{Task Similarity from Noise Variances}
\label{sec:taskclustering}
Our model proposes learning a noise distribution that filters irrelevant dimensions from the latent space for a particular task. From this information-restricting layer we can analyze the latent dimensions each task uses and infer similarities between tasks. Here we investigate the corresponding capabilities of our approach on a synthetic MTL setup derived from MNIST.

To this end we constructed from MNIST a dataset for multi-task classification, where the individual tasks consist of multi-class classification for subsets of MNIST digits. For example, we may consider a task $\mathcal{T}_1=[2,6,0]$ that decides whether the input image $X$ corresponds to labels $Y=2$, $Y=6$, $Y=0$, or none of them ($Y=\varnothing$).  We refer to this setting as Grouped-MNIST.

Our assumption is that tasks with larger overlapping subsets will thus share more information, while tasks that do not share digits will share less. Specifically, we believe that task $\mathcal{T}_2=[2,6,9]$ will share more information with $\mathcal{T}_1=[2,6,0]$ than with $\mathcal{T}_3=[4,8,3]$. This simple setup allows us to evaluate our model for configurations of Grouped-MNIST that differ in the number of groups, the number of digits per group, the number of digits different groups have in common, and the choice of digits per group.
\begin{figure*}[!htbp]
        \centering
        \subfloat{
            \includegraphics[
            trim=0 25 0 10,
            clip,
            width=\three\textwidth]{"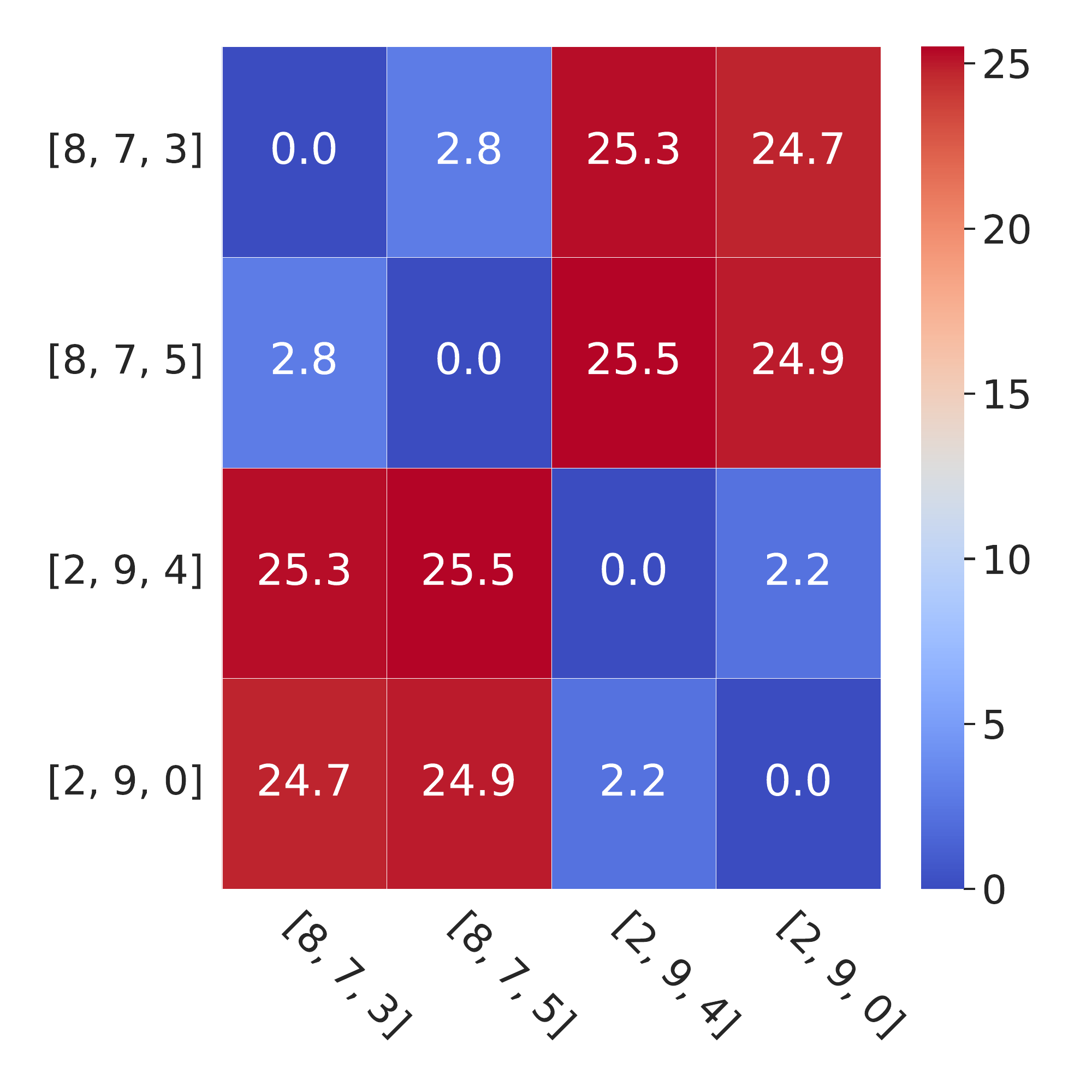"}
        }
        \hfil
        \subfloat{
            \includegraphics[
                trim=0 25 0 10,
                clip,
                width=\three\textwidth]{"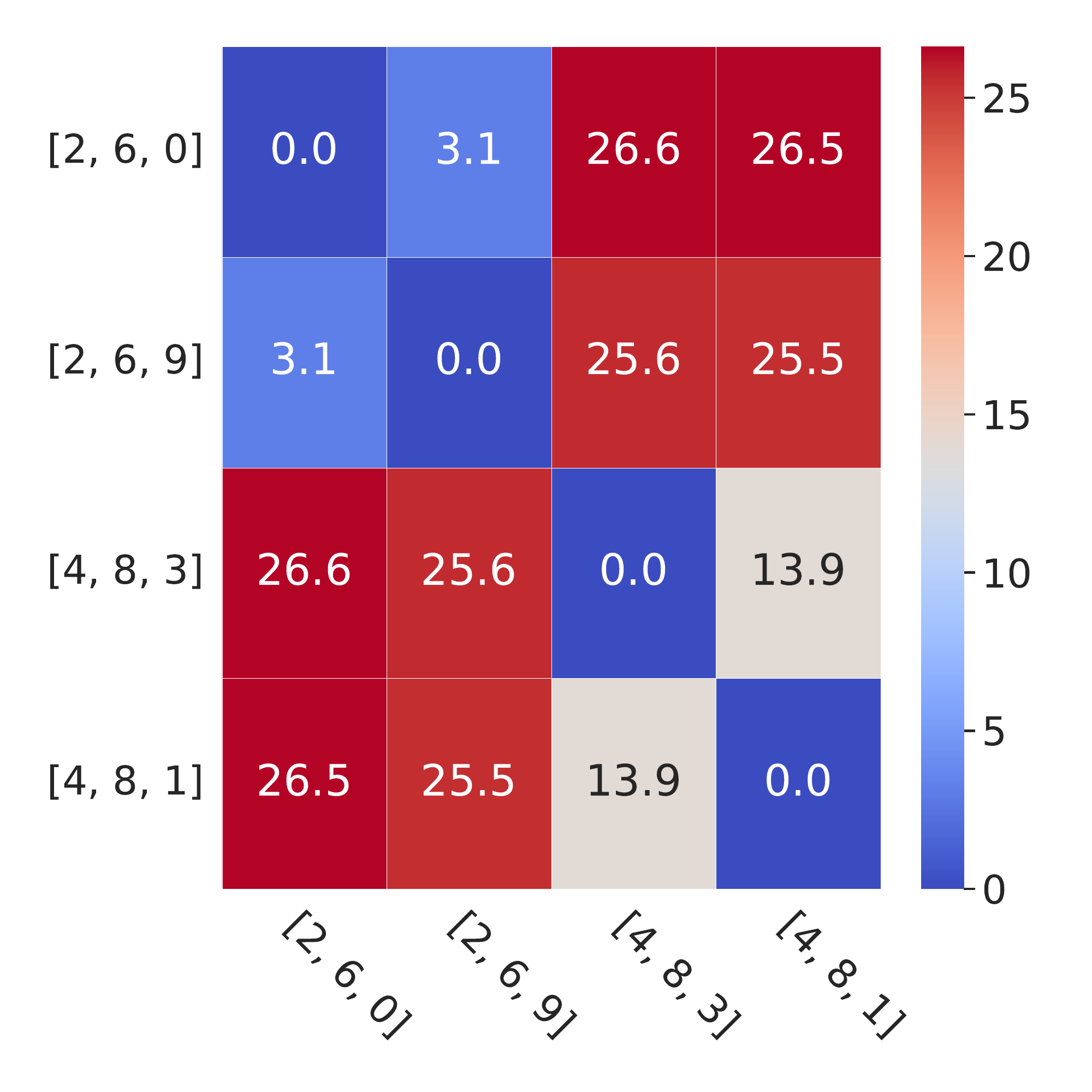"}
        }
        \hfil
        \subfloat{
            \includegraphics[
                trim=0 25 0 10,
                clip,
                width=\three\textwidth]{"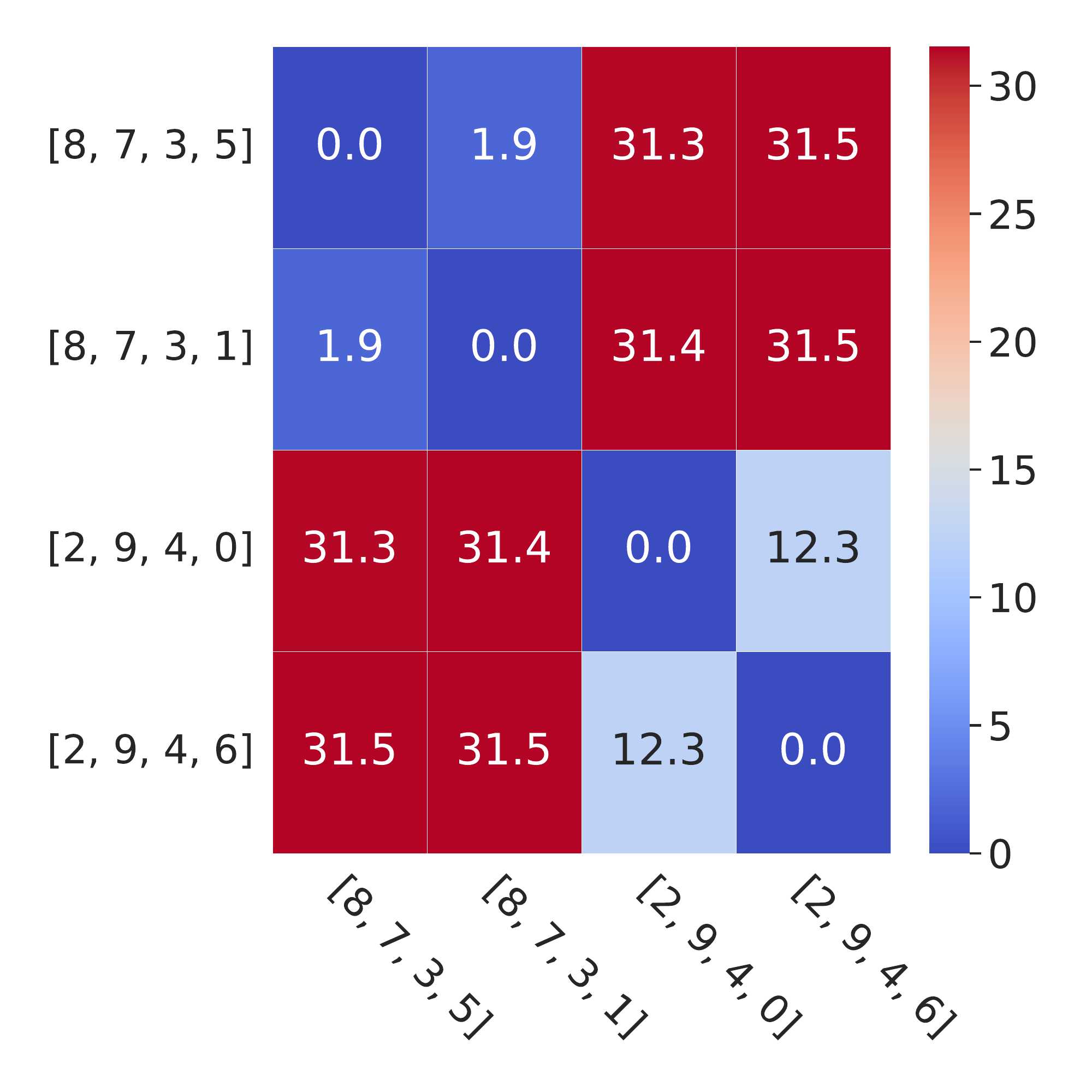"}
        }
        \\
        \subfloat{
             \includegraphics[
                trim=0 48 0 10,
                clip,
                width=\three\textwidth]{"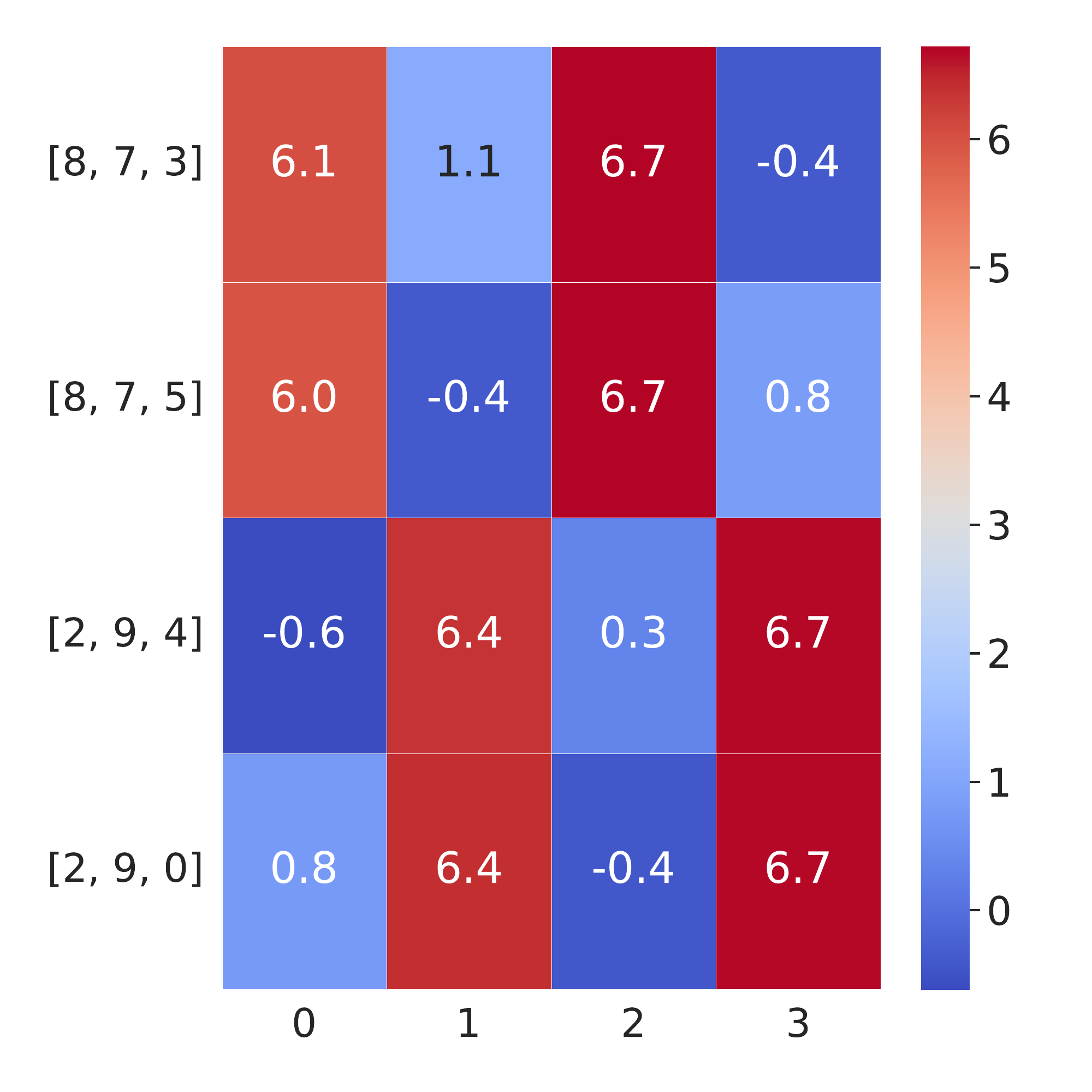"}
        }
        \hfil
        \subfloat{
             \includegraphics[
                trim=0 48 0 10,
                clip,
                width=\three\textwidth]{"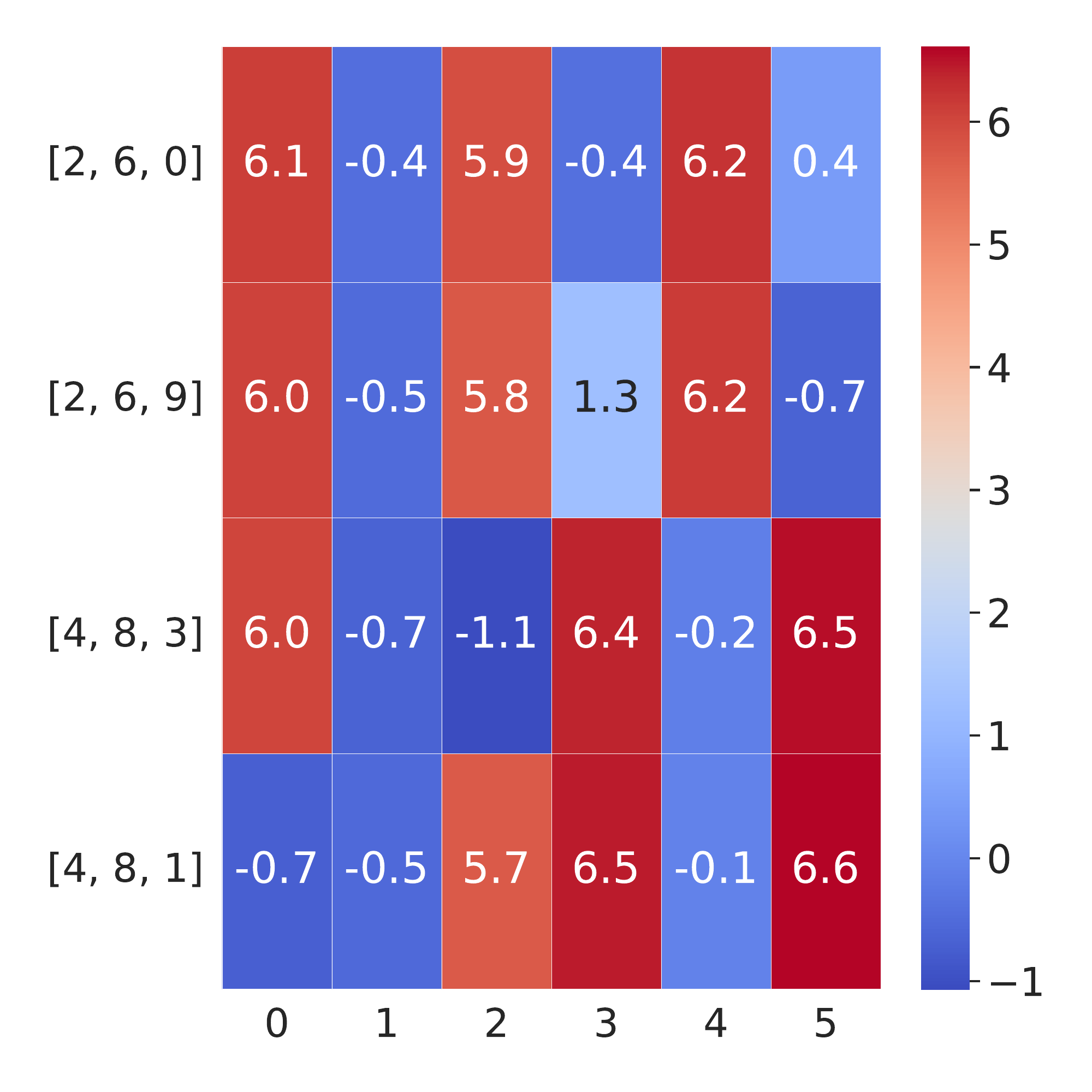"}
        }
        \hfil
        \subfloat{
             \includegraphics[
                trim=0 48 0 10,
                clip,
                width=\three\textwidth]{"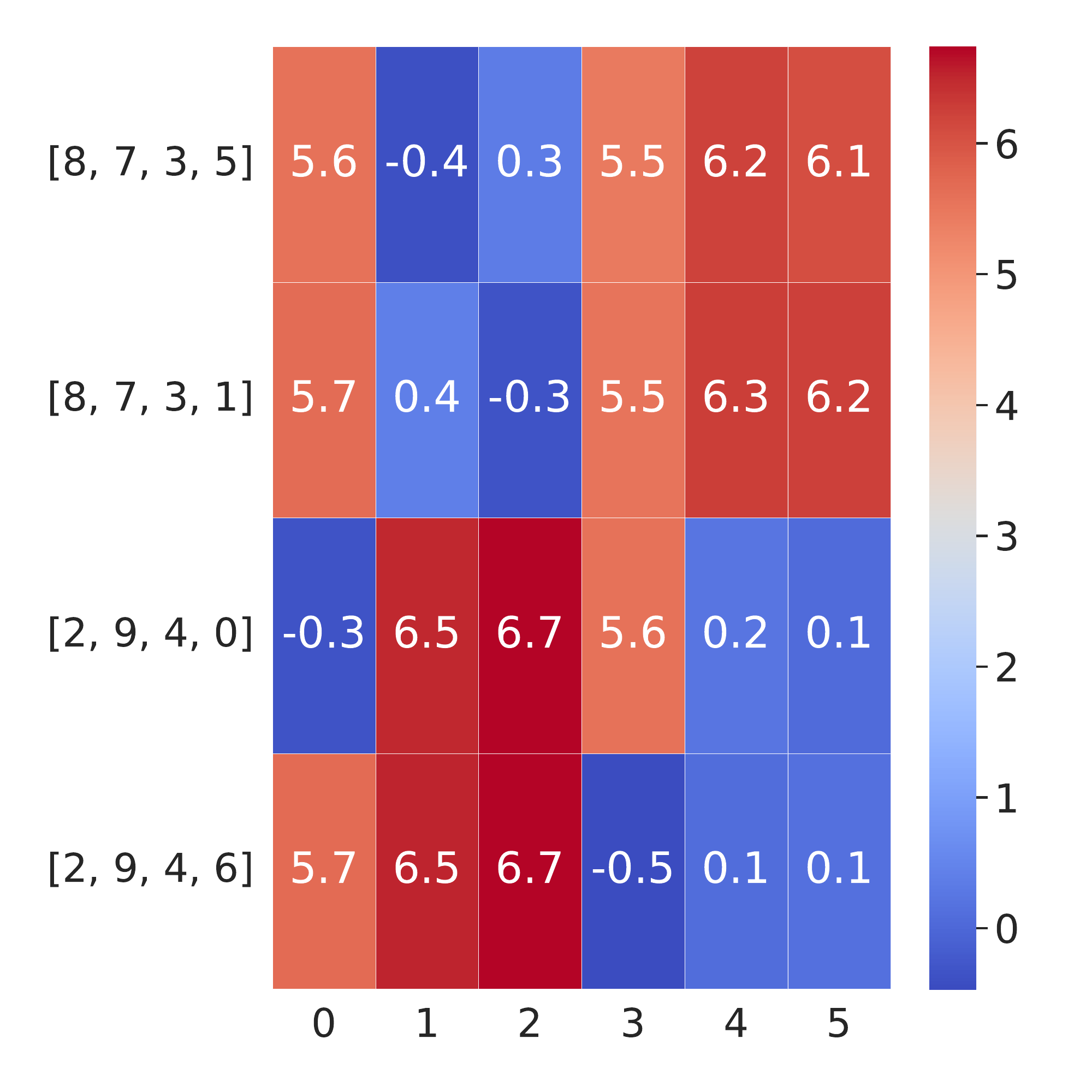"}
        }
        \caption{$L_1$ distance between the noise $\log$-variance for each task (top row) for the Grouped-MNIST dataset with 4 groups of 3 elements each (first and second column) and 4 groups of 4 elements each (third column). In the bottom row, we display the noise $\log$-variances for each task (y-axis). All columns show the results for $\beta=0.1$ with different group configurations, and latent spaces of size 4 and 6. Tasks that share more characteristics have $\log$-variance vectors that are more similar.
        } 
        \label{fig:mnist_pos43}
\end{figure*}
We used this dataset to investigate whether indeed the learned noise variances of our model yield insights about task similarity consistently. Further, in an ablation analysis, we confirmed that the information-restricting layer consisting of additive noise is integral in enabling these insights. To do so, we measure the dissimilarity between two tasks via the $L_1$-distance between the $\log$-variances of the corresponding noise variances. Specifically, the dissimilarity between tasks $\mathcal{T}_1$ and $\mathcal{T}_2$ is given by
\begin{equation}\label{eq:distance}
    \sum_{i=1}^D |\log \sigma_{1,i}^2 - \log \sigma_{2,i}^2| = \sum_{i=1}^D \left| 2\log \frac{\sigma_{1,i}}{\sigma_{2,i}} \right|.
\end{equation}
The network architecture used in this set of experiments is in Table~\ref{tab:mnist_arch}. Unless noted otherwise, all experiments used $\beta=0.1$ and each model was trained for 100 epochs with different random initialization and also using the Adam optimizer~\cite{kingma2017adam} with default parameters and learning rate $10^{-4}$. Again, the weight initialization of the noise variances followed a Gaussian distribution with standard deviation of 0.01. 

\textbf{Results.} \figurename~\ref{fig:mnist_pos43} shows the $L_1$-distance between the noise log-variances of the different tasks (top row) and the noise log-variances for each task (bottom). As anticipated, tasks that share a larger number of digits are more similar to each other than tasks that share less digits. Indeed, looking at the bottom left image of Fig.~\ref{fig:mnist_pos43}, one can see that tasks $[2,9,4]$ and $[2,9,0]$ add noise with low variance to latent dimensions 0 and 2, while the noise added to latent dimensions 1 and 3 is large. I.e., these two tasks rely mostly on information from the first two latent dimensions. In contrast, the two tasks $[8,7,3]$ and $[8,7,5]$ utilize information from mainly latent dimensions 1 and 3, ignoring information from the other two latent dimensions. This leads to the situation depicted in the top left image of Fig.~\ref{fig:mnist_pos43}, where the four tasks are clearly clustered w.r.t.\ the dissimilarity as computed in~\eqref{eq:distance}. A similar picture can be seen for a different group composition and for a different size of the latent space, respectively. 

\begin{figure*}[!htbp]
        \centering
        \subfloat{
            \centering 
            \includegraphics[
                trim=0 25 0 10,
                clip,
                width=\three\textwidth
            ]{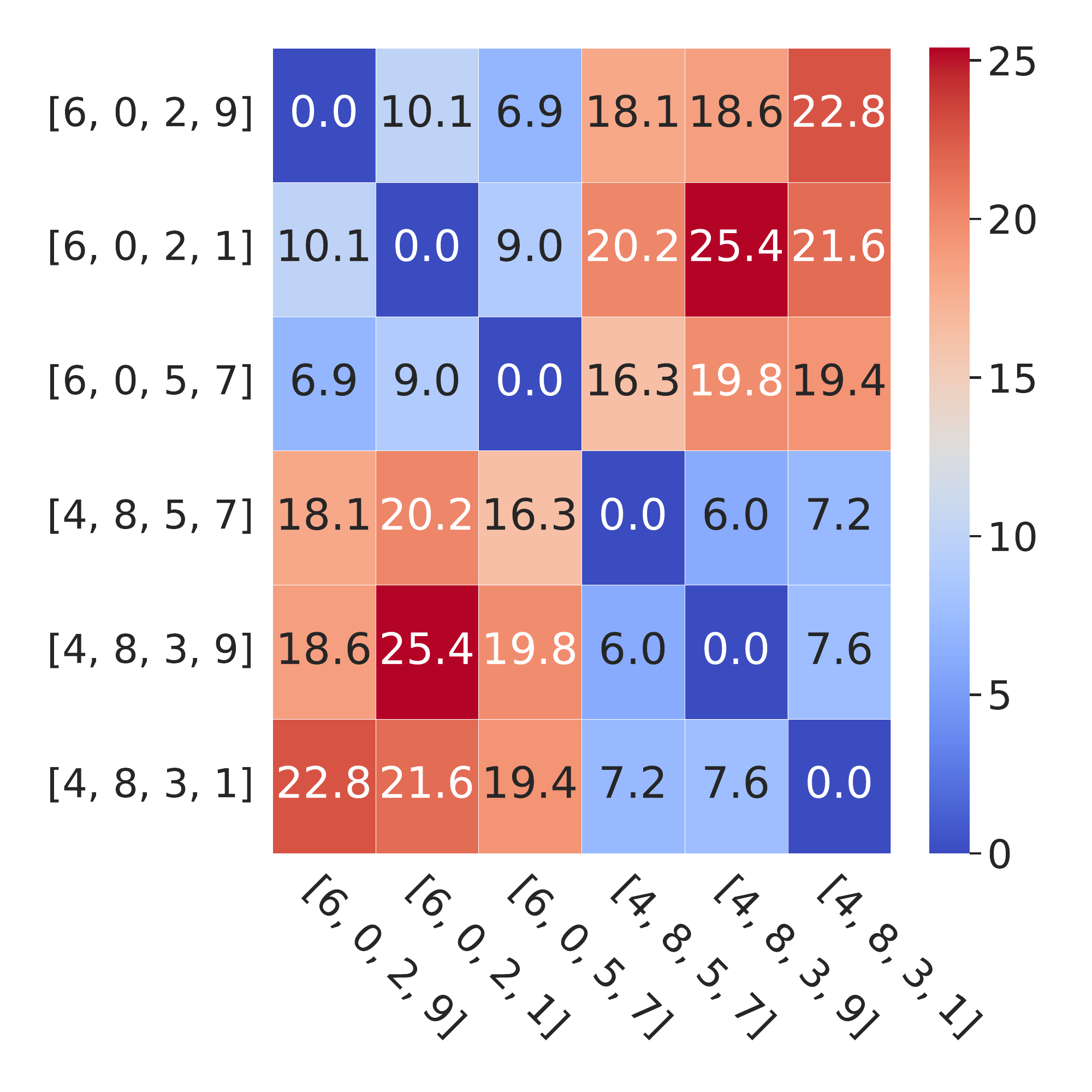}
        }
        \hfil
        \subfloat{
            \centering 
            \includegraphics[
                trim=0 25 0 10,
                clip,
                width=\three\textwidth
            ]{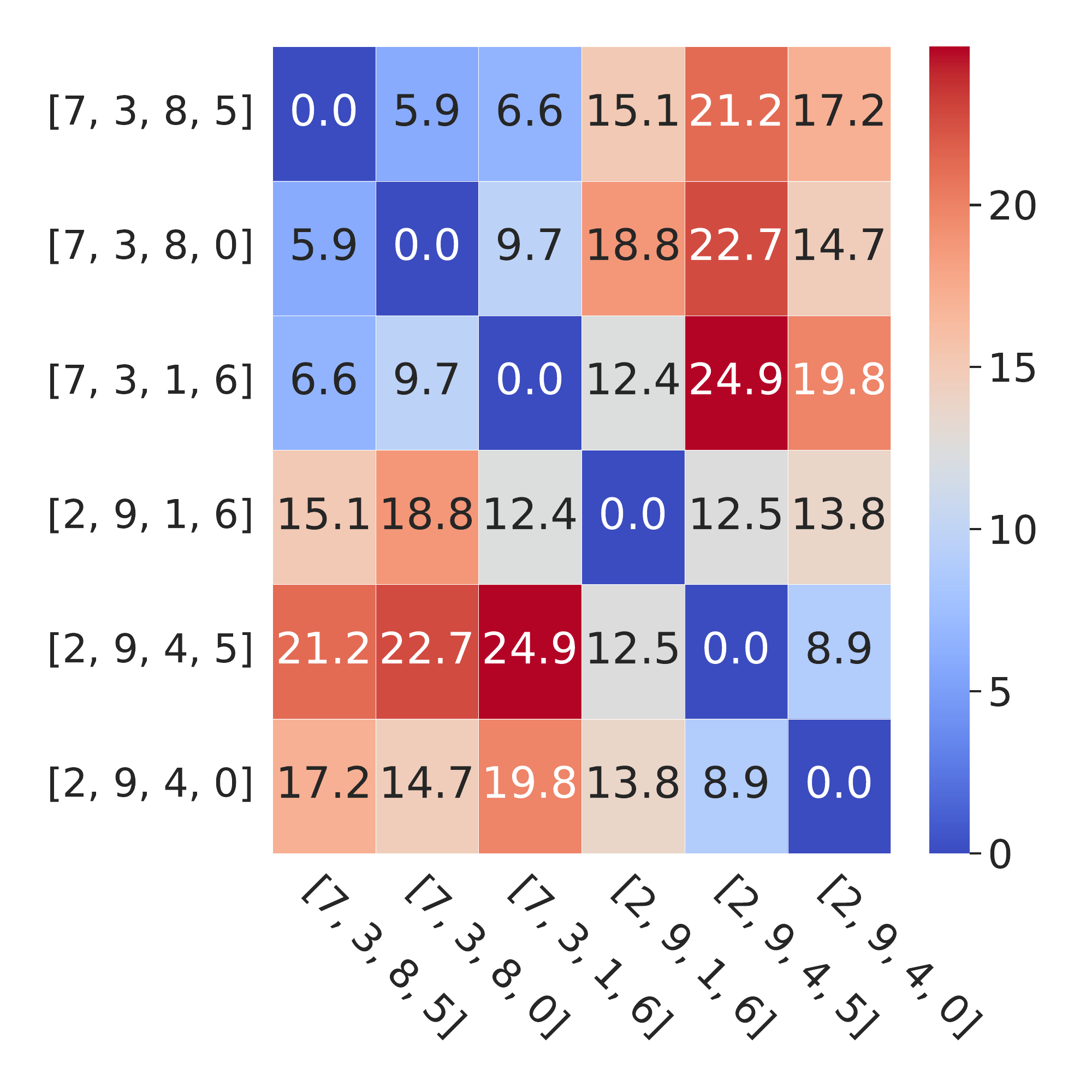}
        }
        \hfil
        \subfloat{
            \centering 
            \includegraphics[
                trim=0 25 0 10,
                clip,
                width=\three\textwidth
            ]{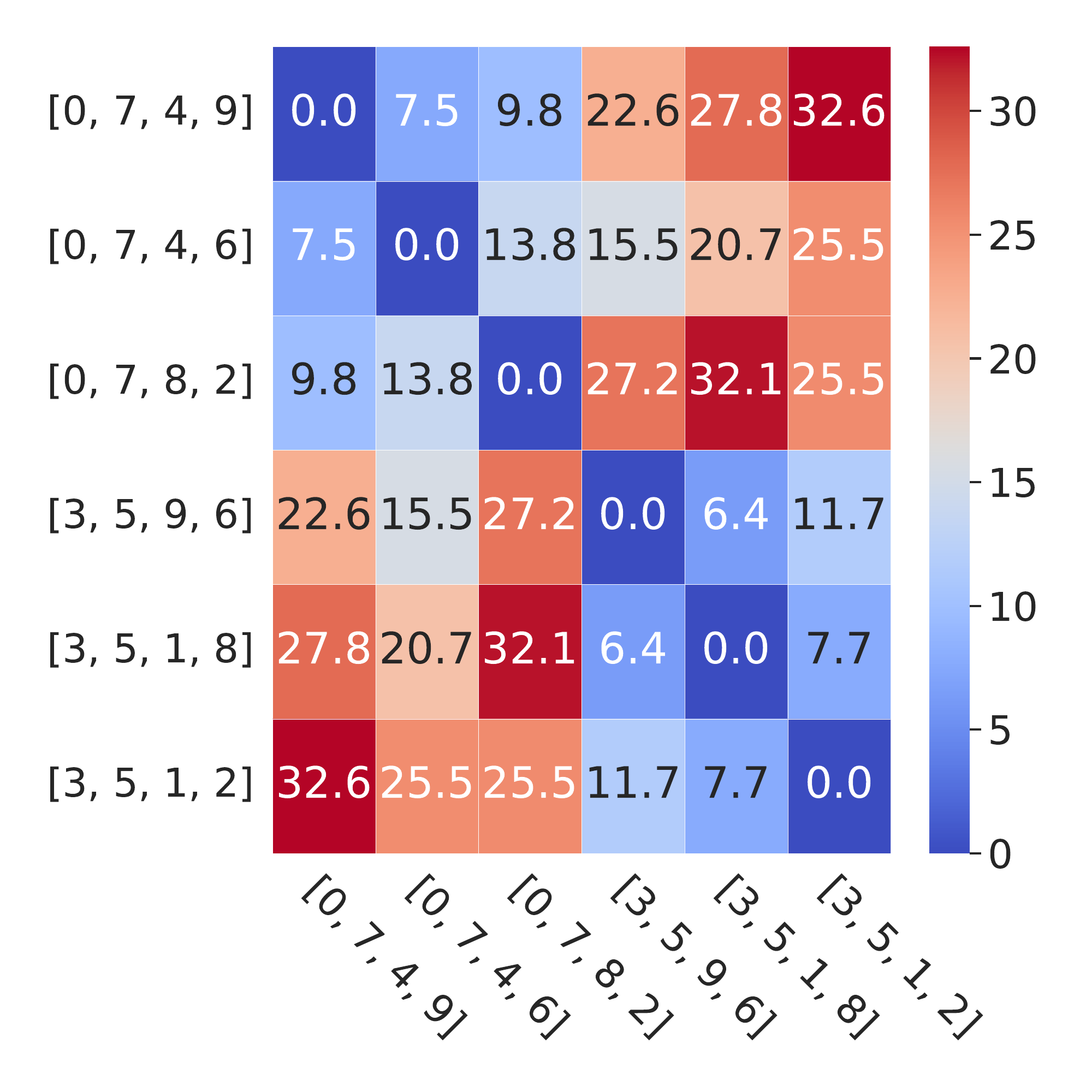}
        }
        \caption{A different configuration of the Grouped-MNIST dataset with more tasks. $L_1$ distance between the noise $\log$-variance for each task. Latent space of size 8, $\beta=0.1$, and three different group configurations.} 
        \label{fig:mnist_pos64}
\end{figure*}

Comparable results can also be seen for a larger number of tasks and different group configurations in Fig.~\ref{fig:mnist_pos64}. In this experiment, we have set different levels of similarity between the tasks. For instance, in the middle figure, task $[7,3,8,0]$ is more similar to task $[7,3,8,5]$ than to task $[2,9,1,6]$. It can be observed that task similarity is still perceptible in most cases, including the less similar tasks. Since our aim in this section was to illustrate how noise variances can inform about task similarity and not about achieving competitive predictive performance, the average task accuracies for the experiment in Fig.~\ref{fig:mnist_pos64} are just around 70\%. This may explain some of it is failures in recognizing the similarity between tasks.

\begin{figure*}[!htbp]
    \centering
    \subfloat[$\beta=0$, $\mathrm{Acc}=84.4\%$]{
        \includegraphics[
                trim=0 25 0 25,
                clip,
                width=\four\textwidth]{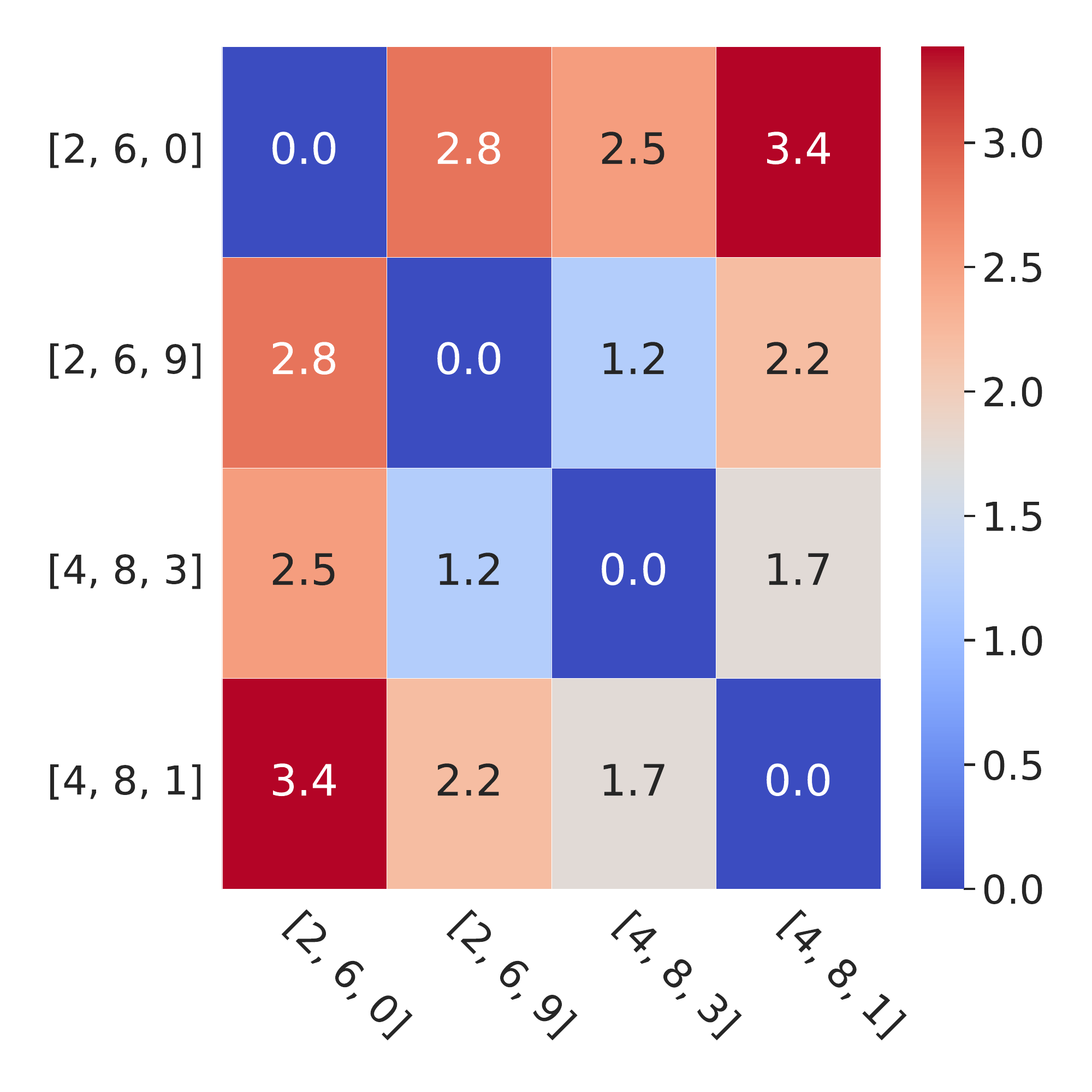}
    }
    \hfil
    \subfloat[\label{fig:beta001}$\beta=0.01$, $\mathrm{Acc} =75.1\%$]{
        \includegraphics[
                trim=0 25 0 25,
                clip,
                width=\four\textwidth]{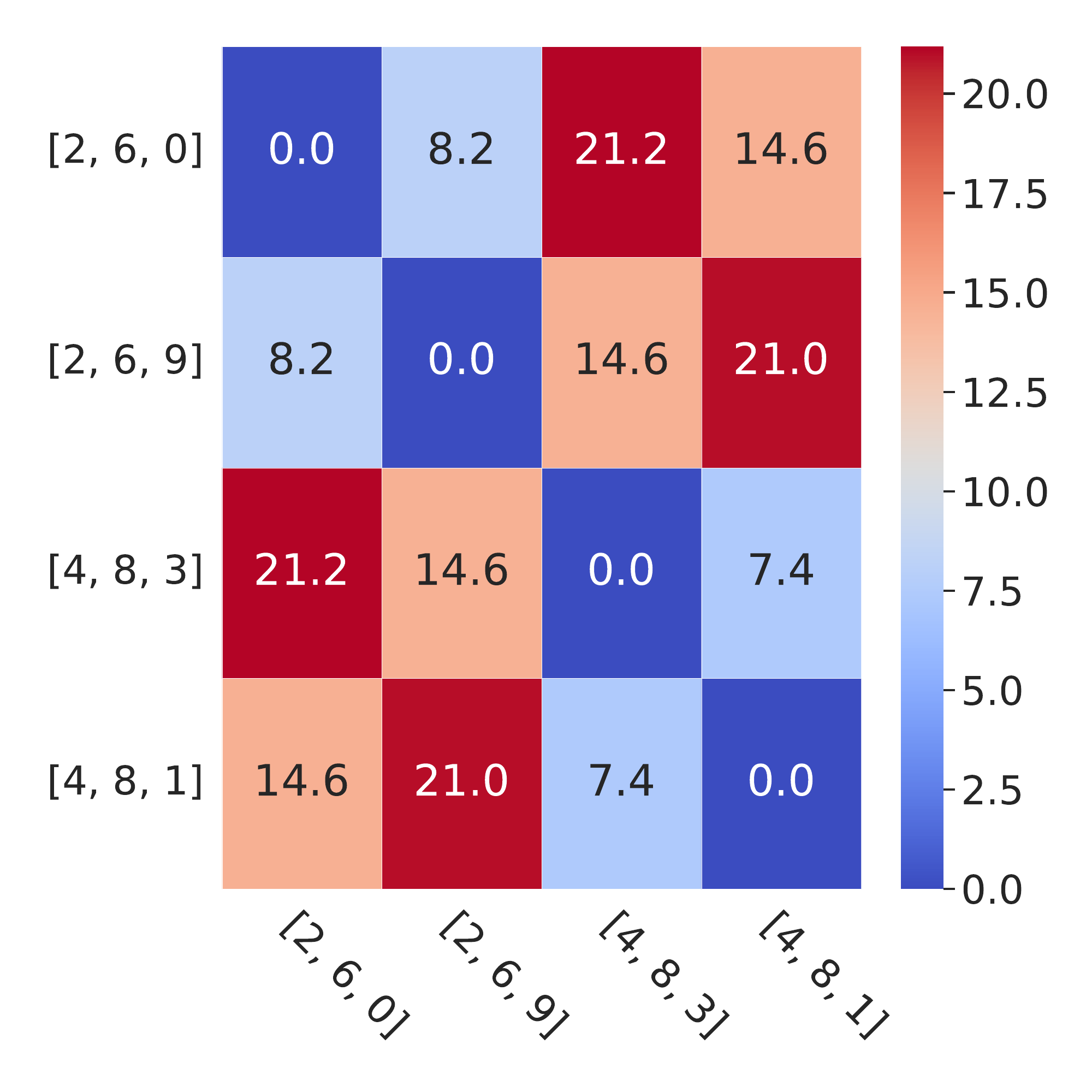}
    }
    \hfil
    \subfloat[\label{fig:beta01}$\beta=0.1$, $\mathrm{Acc} =70.4\%$]{
        \includegraphics[
                trim=0 25 0 25,
                clip,
                width=\four\textwidth]{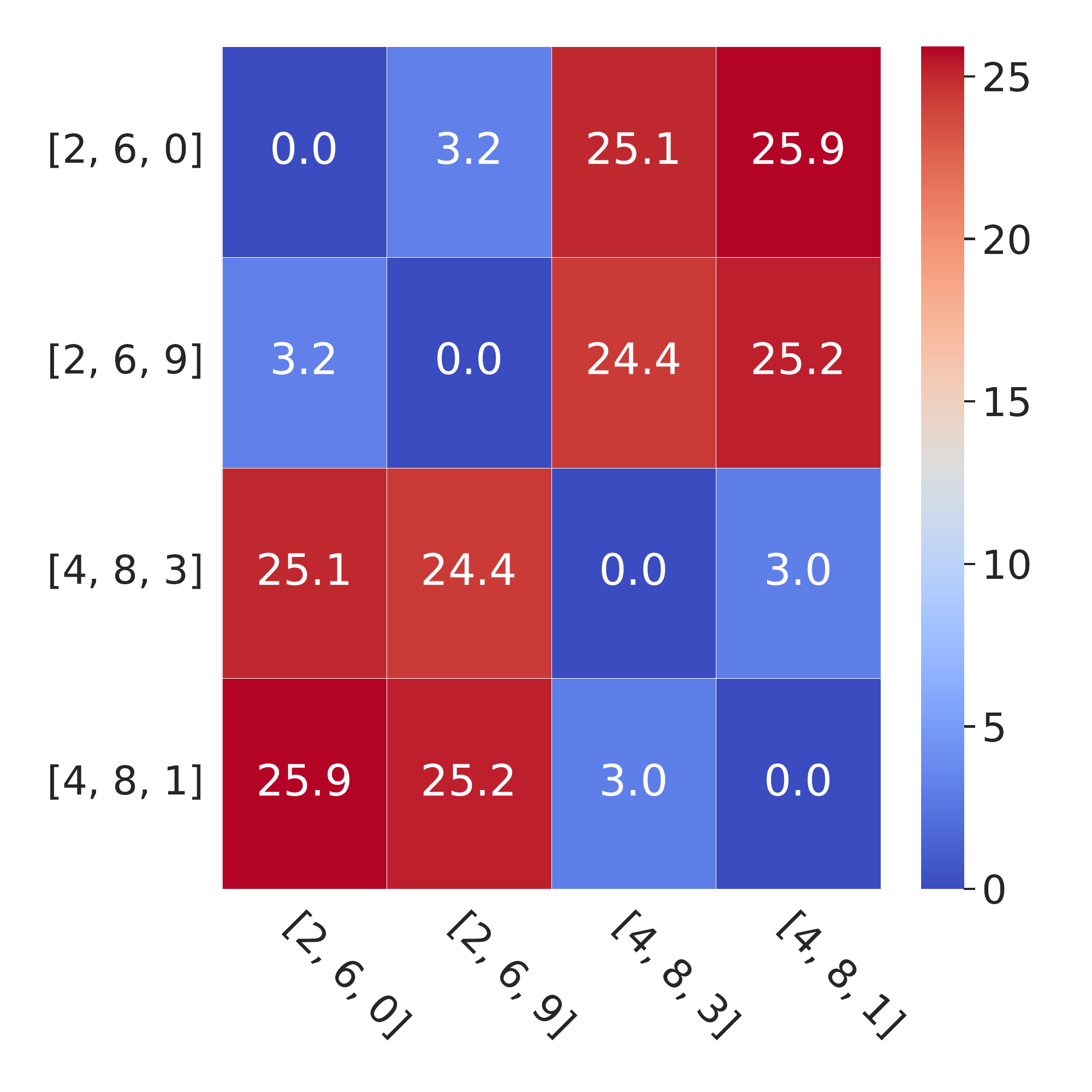}
    }
    \hfil
    \subfloat[$\beta=0.5$, $\mathrm{Acc} =69.8\%$]{
                \includegraphics[
                trim=0 25 0 25,
                clip,
                width=\four\textwidth]{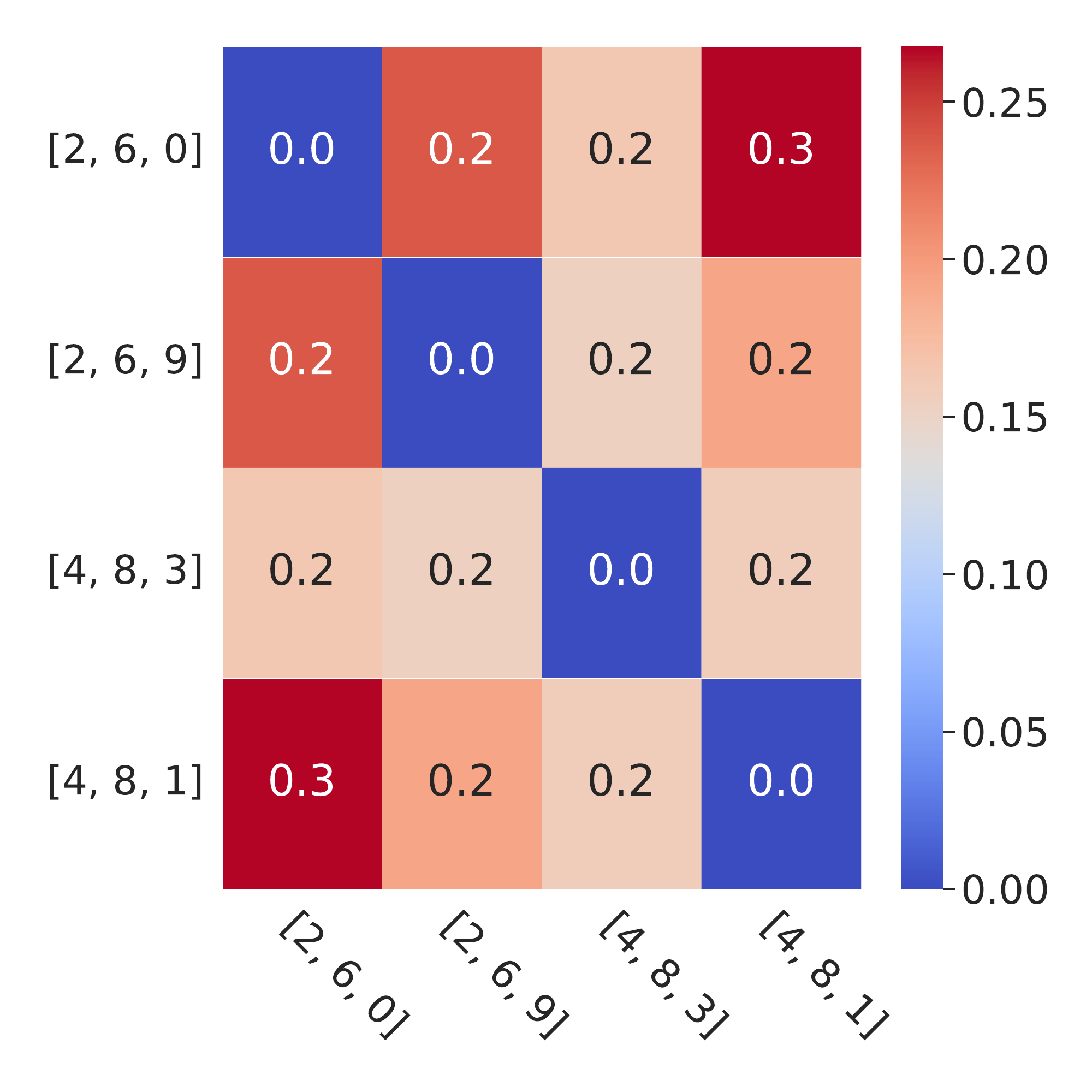}
    }
    \caption{$L_1$ distance between the noise $\log$-variance and average accuracy between tasks for $\beta=0,0.01,0.1,0.5$ (left to right). Latent space of size 4.
    } 
    \label{fig:sens_mnist_pos43}
\end{figure*}

\textbf{Sensitivity Analysis.} While the results in Figs.~\ref{fig:mnist_pos43} and Fig.~\ref{fig:mnist_pos64} have been obtained for $\beta=0.1$, here we investigate the effect of $\beta$ on the detection of task similarity. To this end, again for the simple architecture of Table~\ref{tab:mnist_arch}, we varied $\beta$ between 0 and 0.5 (Fig.~\ref{fig:sens_mnist_pos43}).
As already observed in Section~\ref{sec:taskclustering}, increasing values of $\beta$ correspond to a decreasing predictive performance. However, we also observe that attaining high similarity between information filters is sensitive to the level of compression (regularization) set by the $\beta$ parameter. I.e., for both too small and too high values of $\beta$, task similarity is not well represented by the corresponding tasks' log-variances.
In fact, we only observe task clustering in \figurename~\ref{fig:beta001} and \figurename~\ref{fig:beta01}, with a drop of at least 9.3\% in average accuracy.
We recognize that appropriately setting $\beta$ might be challenging for more complex datasets. We leave this and the selection of tasks-specific $\beta_j$ for future work.

\textbf{Ablation Analysis.} Our results indicate that the proposed system architecture induces, when parameterized appropriately, a task-specific disentanglement in latent space. We now perform an ablation study to show that our system architecture is indeed instrumental in achieving this disentanglement. To this end, compare the latent representation $Z$ obtained by our approach to one in which the variance of the noise is not trainable but set to $\sigma_{j,i}^2=\mathrm{e}$ for all $j$ and $i$.

Since we measure task similarity and, thus, disentanglement of $Z$ via the variance of noise vectors, we proceed as follows for the setting with fixed noise. In a first step, we train the parameters $\theta$ and $\psi$ of encoder and all classifiers, respectively. Then, we freeze the encoder parameters $\theta$, make the noise variances $\{\sigma_{j,i}^2\}$ trainable, and retrain the classifier parameters $\psi$ jointly with these noise variances. 
 
For both our original system architecture and the one without added noise, we perform experiments with Grouped-MNIST. In both settings, the latent dimension is set to $D=4$. 
The tasks are grouped in pairs, where each pair shares two digits (cf. Fig.~\ref{fig:mnist_pos43}). Finally, the experiment was repeated 50 times with a different choice of digits per group. We trained each trial for 25 epochs. We call disentanglement of $Z$ successful if the shortest distance~\eqref{eq:distance} between tasks from different pairs is larger than the largest distance between tasks from the same pair.
From the results of the experiment seen in Table \ref{tab:repetead_exp}, we can notice that,  for the right choice of $\beta$, our model successfully detects the task pairs much more reliably than the model with fixed noise variances. The success rate of clustering was however significantly high in all cases, comparing this to clustering at random. 
This indicates that the encoder can, even without the learnable filter, construct a task-specifically disentangled latent space. However, disentanglement is improved and strengthened by adding the proposed information filters. Further, information filters are a natural way to determine this disentanglement and, if parameterized appropriately, can improve generalization performance via encouraging compressed representations.
\begin{table}[!htp]
\renewcommand{\arraystretch}{1.3}
\caption{Ablation Analysis Results}
\label{tab:repetead_exp}
\centering
\begin{tabular}{ccc} 
\hline
\bfseries Model &  $\pmb \beta$ & \bfseries Successes (\%)\\
\hline
\multirow{3}{5em}{\centering Ours} & 0.1 & 96\\ 
& 0.05 & 88\\ 
& 0.01 & 68\\ 
\hline
\multirow{3}{5em}{Fixed Noise} & 0.1 & 32\\ 
& 0.05 & 40\\ 
& 0.01 & 56\\ 
\hline
\end{tabular} 
\end{table}
\section{Discussion, Limitations, and Outlook}
\label{sec:discussion}
% Dimensions individually not interpretable
Our approach of quantifying task similarity relies on the fact that the latent representations can be disentangled to some extent: While we do not require that each individual dimension of $Z$ to be connected to a human-interpretable concept, we assume that some dimensions of $Z$ are more useful for a specific task than others. In our synthetic toy example, based on multiple tasks for the MNIST dataset, we were able to show that the learned $Z$ indeed can be split into such subsets of dimensions, which subsequently allows assessing task similarity based on the noise variances defining the task-specific representations $W_j$. However, an interpretation of individual dimensions has so far eluded us, and will even more so if these synthetic toy tasks are replaced by realistic tasks on a real-world dataset.

We have further observed that this kind of disentanglement is very sensitive to parameter settings (which resonates with the general feebleness of disentanglement~\cite{locatello2019challenging}). Specifically, we have observed that, even in synthetic toy examples, architecture and parameterization must be chosen carefully to guarantee that task clustering emerges consistently. Also, the composition of the set of tasks plays a fundamental role in this. We believe that these insights also apply to the results of~\cite{nautrup2020operationally}; i.e., we believe that the learned latent representations of a set of reference experiments are only physically meaningful if the representations are trained with a well-chosen and sufficiently diverse set of experimental predictions (which corresponds to different tasks in our setting).

In this work, we have focused on homogeneous-feature MTL, i.e., the setting where each of the tasks relies on the same set of features, and where we can assume that the marginal distribution of features does not change between tasks. While homogeneous-feature MTL is certainly of practical relevance, the extension of our hierarchical model to heterogeneous-feature MTL~\cite{Zhang_Survey} shall be the focus of future work. Specifically, in settings where different tasks rely on different sets of features or induce different feature distributions, it may be practical to derive task-agnostic representations useful for all tasks.

Finally, while our formulation~\eqref{eq:cost_function} of the problem permits different compression constraints $\beta_j$ for different tasks, in this work, we did not investigate how these hyperparameters are best set but rather studied how $\beta_j=\beta$ influences MTL performance. Neither did we investigate weighing the cross-entropy terms in~\eqref{eq:cost_function} differently for each task, capturing that some tasks may be more difficult to learn than others. In future work, we may thus extend our approach by automatic \emph{uncertainty weighting} as proposed by~\cite{Kendall_2018_CVPR} and utilized in~\cite{qian2021mvib}.

\section{Conclusion}
We proposed a multi-task learning model that restricts information individual tasks can access from a task-agnostic latent representation. By analyzing the information flow from this task-agnostic representation to the task-specific decoders, we can better understand the information each task uses. This yields the opportunity to analyze task similarities by information shared between tasks and, eventually, use this information to combine tasks for further multi-task learning.

\section*{Acknowledgements}
The authors gratefully acknowledge the financial support under the scope of the COMET program within the K2 Center ``Integrated Computational Material, Process and Product Engineering (IC-MPPE)” (Project No 886385) and the COMET Module ``DDAI''. The COMET – Competence Centers for Excellent Technologies program is supported by the Austrian Federal Ministries for Climate Action, Environment, Energy, Mobility, Innovation and Technology (BMK) and for Digital and Economic Affairs (BMDW), represented by the Austrian Research Promotion Agency (FFG), the federal states of Styria, Upper Austria and Tyrol, and by partners from industry and academia. The COMET Programme is managed by FFG.

\bibliographystyle{IEEEtran}
\bibliography{IEEEabrv,references}

\end{document}